\pdfoutput=1

\documentclass[11pt]{article}

\usepackage{acl}

\usepackage{times}
\usepackage{latexsym}

\usepackage[T1]{fontenc}

\usepackage[utf8]{inputenc}

\usepackage{microtype}

\usepackage{inconsolata}

\usepackage{graphicx}

\title{CogMG: Collaborative Augmentation Between Large Language Model and Knowledge Graph}

\author{
Tong Zhou$^1$\and
Yubo Chen$^{1,2}$$^*$\and
Kang Liu$^{1,2,3}$\and
Jun Zhao$^{1,2}$$^*$\\
$^1$The Laboratory of Cognition and Decision Intelligence for Complex Systems \\
Institute of Automation, Chinese Academy of Sciences\\
$^2$School of Artificial Intelligence, University of Chinese Academy of Sciences\\
$^3$Shanghai Artificial Intelligence Laboratory  \\
tong.zhou@ia.ac.cn,
\{yubo.chen, kliu, jzhao\}@nlpr.ia.ac.cn
}

\begin{document}

\maketitle

\begin{abstract}
Large language models have become integral to question-answering applications despite their propensity for generating hallucinations and factually inaccurate content. 
Querying knowledge graphs to reduce hallucinations in LLM meets the challenge of incomplete knowledge coverage in knowledge graphs. 
On the other hand, updating knowledge graphs by information extraction and knowledge graph completion faces the knowledge update misalignment issue.
In this work, we introduce a collaborative augmentation framework, CogMG, leveraging knowledge graphs to address the limitations of LLMs in QA scenarios, explicitly targeting the problems of incomplete knowledge coverage and knowledge update misalignment. 
The LLMs identify and decompose required knowledge triples that are not present in the KG, enriching them and aligning updates with real-world demands. 
We demonstrate the efficacy of this approach through a supervised fine-tuned LLM within an agent framework, showing significant improvements in reducing hallucinations and enhancing factual accuracy in QA responses. 
Our code\footnotemark[1] and video\footnotemark[2] are publicly available.
\end{abstract}

\renewcommand*{\thefootnote}{\fnsymbol{footnote}} 
\footnotetext[1]{Corresponding author}

\begingroup
\renewcommand*{\thefootnote}{\arabic{footnote}} 
\setcounter{footnote}{0}
\footnotetext[1]{Project: \href{https://github.com/tongzhou21/CogMG}{https://github.com/tongzhou21/CogMG}} 
\footnotetext[2]{Video: \href{https://youtu.be/WnkS0Qk_0OM}{https://youtu.be/WnkS0Qk\_0OM}} 
\endgroup

\section{Introduction}

Large language models (LLMs) \citep{brown2020language,achiam2023gpt} have witnessed a surge in adoption for question-answering (QA) applications \citep{stelmakh2022asqa}. 
Despite their ability to produce engaging and coherent responses, these models are susceptible to generating hallucinated content and frequently encompass factually inaccurate information \citep{rawte2023survey}.
\citealp{xu2024hallucination} indicated that this inevitable symptom imputes their data \citep{kandpal2023large}, training \citep{liu2024exposing}, and inference stages \citep{dziri2021neural}.
Fortunately, LLMs can leverage their comprehension and reasoning ability by referring to external knowledge sources to relieve hallucinations, such as documents \citep{lewis2020retrieval} and knowledge graphs \citep{sun2023think}.
We concentrate on utilizing knowledge graphs (KGs), which provide a complementary strength to Large Language Models (LLMs) through their structured format and precise encapsulation of factual information. 
However, the utility of KGs in QA scenarios is hindered by the challenges of \textit{incomplete knowledge coverage} and \textit{knowledge update misalignment}.

\textbf{Incomplete Knowledge Coverage:} 
In principle, knowledge graphs possess the capability to encompass a vast array of information; however, they are also confronted with the challenge of achieving comprehensive coverage in their storage of knowledge.
The explicitly encoded triples within the KG prove inadequate to exhaustively cover the knowledge required for practical QA scenarios.
Existing approaches to augmenting QA systems with KG have primarily focused on improving parsing formal language \citep{xiong2024interactive} or semantic relevance in retrieval knowledge triples \citep{wu2023retrieve}, pursuing the corresponding knowledge pre-storage in the KG for pre-defined questions.
There is relatively limited attention given to the subsequent handling of queries that do not hit the knowledge graph.

\textbf{Knowledge Update Misalignment:} 
Current approaches to updating knowledge graphs primarily depend on two strategies: extracting knowledge triples from unstructured text \citep{wang2023instructuie,xiao2023yayi} (Information Extraction) and inferring unseen linkages through the analysis of existing connections between nodes \citep{yang2023cp} (Knowledge Graph Completion). 
These paradigms employed for updating KGs are characterized by their aimless and seemingly infinite nature and, therefore, do not fully address the misalignment between the newly acquired knowledge and real-world user needs.
This highlights a lack of proactive consideration in updating the knowledge graph to align better with user demands.

\begin{figure*} 
  \centering
  \includegraphics[width=0.99\textwidth]{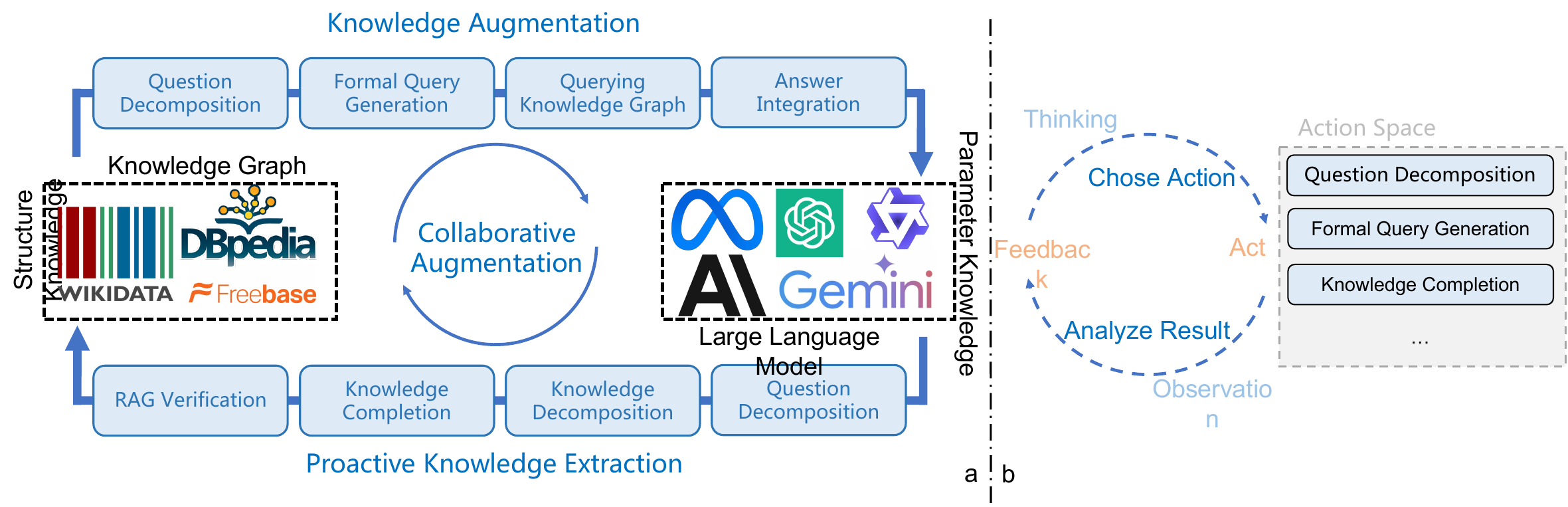}
  \caption{Left part: A schematic diagram illustrating the overall design of the collaborative augmentation framework CogMG, involving LLM and KG. Right part: We implement CogMG using an agent-based framework, with each module designed to be plug-and-play to ensure generalizability.}
  \label{figure_ar_2}
\end{figure*}

To address the above two challenges, this paper proposes a framework called CogMG for collaborative augmentation between LLM and KG.
When a query exceeds the knowledge scope of the current KG, the LLM is encouraged to explicitly decompose the required knowledge triples.
Subsequently, completion is done based on the extensive knowledge encoded in the LLM's parameters, serving as the reference for the final answer.
The explicit identification of necessary knowledge triples serves as a means for model introspection to mitigate hallucination and proactively highlights deficiencies in the KG in meeting real-world demands.
Moreover, identifying these triples allows for their automatic verification through retrieval augmented generation (RAG) with external documents.
The retrieved relevant documents can also be a reference for manual review before incorporating triples into the knowledge graph.
This continual and proactive process of knowledge updating enables the knowledge graph to meet actual knowledge demands gradually. 
Consequently, the LLM can leverage the augmented KG to improve its factualness in answering questions, forming a collaborative augmentation between LLM and KG.
The main contributions of this paper are shown below:
\begin{itemize}
\item We propose the collaborative augmentation framework between LLM and KG, which is called CogMG.  Address knowledge deficiency in LLMs and advocate actively updating the knowledge within the KG according to user demand.
\item We fine-tune an open-source LLM to adapt the collaborative augmentation paradigm CogMG in an agent framework and demonstrate it by implementing a website system. The agent framework is modular and pluggable, and the system is interactive and user-friendly.
\item According to a use-case presentation and the experimental results in various situations, we demonstrate the effectiveness of CogMG in updating knowledge proactively and enhancing response quality in various real-world QA scenarios.
\end{itemize}

\section{Framework Design}

The single iteration of CogMG framework comprises three steps: (1) Querying the Knowledge Graph: Large models utilize reasoning and planning capabilities to decompose queries and generate formalized query statements for querying the knowledge graph. (2) Processing Results: If results are returned successfully, detailed answers preferred by humans are integrated. If unsuccessful, the required triples are explicitly identified and broken down before being integrated into the answer. (3) Graph Evolution: Utilizing external knowledge verification and modification to incorporate triples that were not hit into the knowledge graph.

\subsection{Querying Knowledge Graph}

Given a knowledge-intensive question, we initiate our approach by deconstructing the corresponding formal query into sub-steps in natural language. 
This decomposition aids in elucidating the necessary and universal logic for querying knowledge graphs, ensuring our method's generalizability across various KG schemas. 
The LLM then calls a formal language parsing tool to execute the query. 
This tool receives the logically decomposed steps in natural language as input, translates them into a formal query language tailored to the target knowledge graph, and returns the query results.

\subsection{Processing Result}

Upon receiving the query results from KG, the LLM leverages its comprehension and reasoning capabilities to organize the final answer. 
If the query execution encounters errors, the LLM delineates the essential knowledge triples with unknown components based on decomposed steps. 
Suppose the complement of these triples could provide the necessary knowledge to answer the question.
Subsequently, knowledge encoded within the model's parameters is utilized to complete these triples. And then, the model generates the final answer according to these facts. Note that the completion step is applicable to LLMs with capabilities of any level. 
Explicit the necessary knowledge not only mitigates the hallucination effect due to snowballing in the current output but also identifies knowledge gaps within the graph, thereby facilitating the enhancement of the graph's knowledge coverage. 
The incomplete knowledge triples, and their completions are logged for potential incorporation into the graph or further verification.

\subsection{Knowledge Graph Evolution}

The high generality and broad coverage of parameter knowledge encoded within LLM can supplement the more specialized knowledge in KG.
These triples completed by LLM can be added to KG directly.
However, the LLM struggles with rare, long-tail, and domain-specific knowledge and lacks robustness in its knowledge statement.
We offer an option for manual intervention, where administrators can choose to (1) directly incorporate the completed triples into the knowledge graph, (2) manually adjust them before addition, or (3) verify them automatically according to external knowledge sources.

To automatically validate and correct these triples, CogMG searches related documents within unstructured corpora and makes comparisons in facts between documents and triples.
These documents, which could drawn from domain-specific texts, general encyclopedias, or rapidly updated search engines, not only enhance the factual accuracy of the knowledge but also provide interpretable references for manual review.
Based on the insights from these external sources, the model adjusts the proposed knowledge triples, making them suitable for manual inclusion into the knowledge graph.


\section{Implementation and Usecase}

We fine-tune an open-source LLM for implementing the CogMG and develop an online system to demonstrate and evaluate our proposed collaborative augmentation framework.

\subsection{Model and Components}

LLM is capable of serving as an agent to plan and call tools independently \citep{qin2023toolllm}. 
We adopted ReAct's agent framework \citep{yao2022react} to adapt LLM to our proposition of modularization and generalization in CogMG's philosophy. 
We utilize a subset of Wikidata as the knowledge graph, KoPL \citep{cao2022kqa} as the query engine, and the KQA-Pro dataset as the backbone of the fine-tuning dataset.
KQA-Pro contains natural language questions with corresponding KoPL queries, SPARQL queries, and the gold answer.
To ensure that the agent exhibits the expected behavior across various scenarios, we construct customized SFT datasets to fine-tune or utilize in-context learning to prompt the model. 
Qwen-14B-Chat \citep{bai2023qwen} is responsible for all the SFT data generation and the agent backbone.
Notably, our framework is applicable across various knowledge graphs and LLM.
We will introduce our solution scenario by scenario.

\textbf{Question Decomposition:} Utilizing decomposition steps as intermediaries between questions and formal languages clarifies problem-solving logic and enhances robustness against different expressions of questions.
We manually write several natural language explanations of query steps to obtain parallel data cases between questions and natural language explanations according to corresponding KoPL function calls. 
With these parallel examples, we prompt an LLM to generate sub-steps across the entire dataset and get 50k pairs of data.
These data are reserved to construct agent behavioral SFT dataset.

\begin{figure*} 
  \centering
  \includegraphics[width=1.0\textwidth]{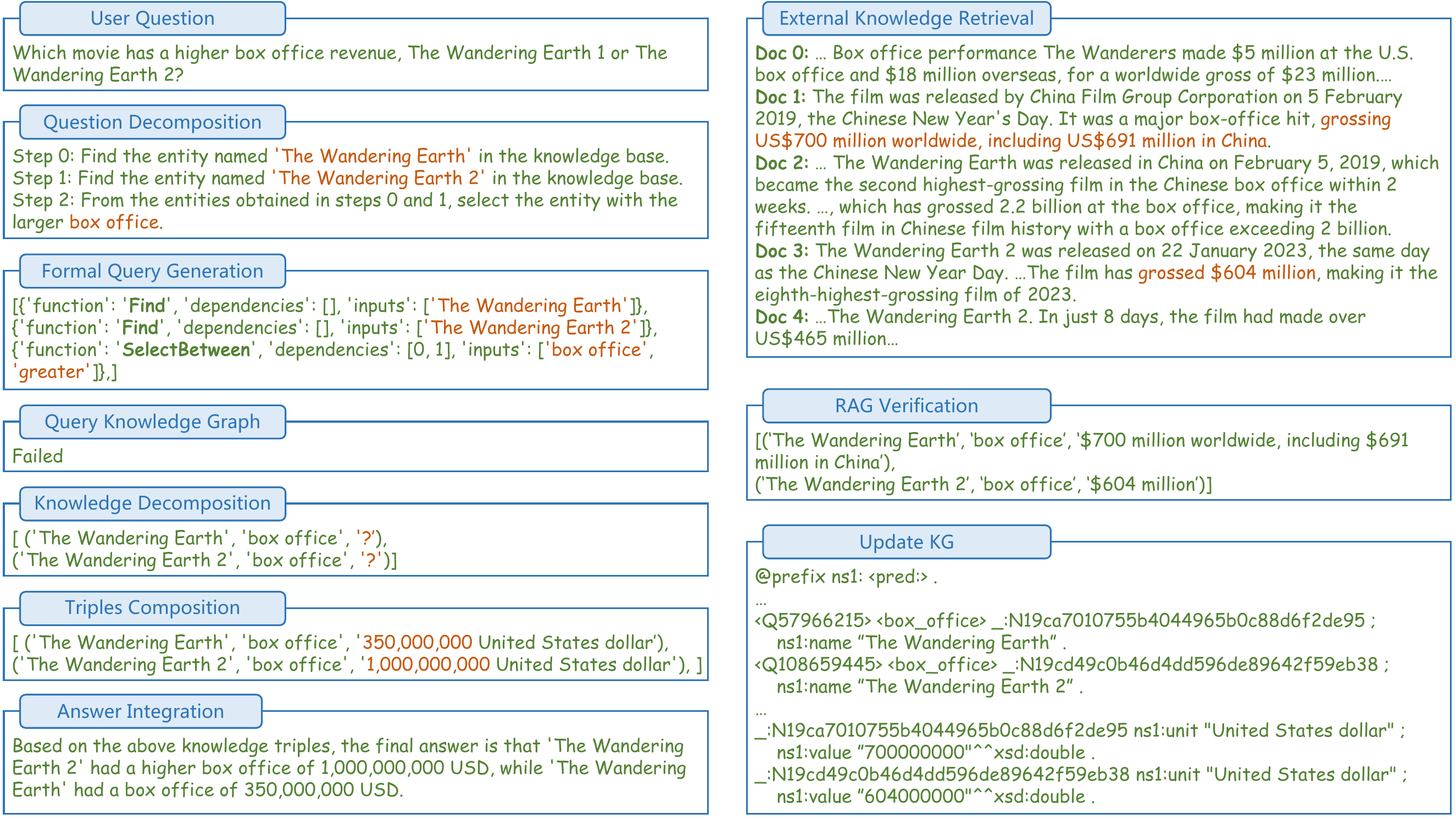}
  \caption{A complete example of how the system handles queries not found in the knowledge graph and the processing of related knowledge.}
  \label{figure_case_0}
\end{figure*}

\textbf{Formal Query Generation:} A model that parses the natural language explanations of query steps to the KoPL formal program could be rapidly trained using the parallel data.
Since the parsing process is relatively undemanding on the model's capabilities, we fine-tuned a 7B model to create a dedicated model in the tool of querying knowledge graph.

\textbf{Querying Knowledge Graph:} We wrapped the execution of the KoPL engine to uniformly return "Failed" upon errors, facilitating the model's decision-making and recognition.
The query tool processes decomposed step inputs through a parsing model predicts the KoPL query program and returns the results of the knowledge graph query.

\textbf{Answer Integration:} The gold answers provided by KQA Pro are brief and precise at the word level and have gaps with the more detailed explanations preferred by humans.
Hence, we supply the inference model with questions and gold answers from KG execution, instructing it to generate more exhaustive, explanatory responses to each question in the dataset. 
The answer integration scenario is a part of the agent behavior.

\textbf{Knowledge Decomposition:} We explicitly decompose the formal query's target triples to clarify the facts necessary for answering questions.
This step is essential for manually annotating some query statements to incomplete triple, with unknown parts of facts expressed as question marks, and then using these samples as examples for the model to infer the triple decomposition for all data.
Given the precise label names in the KoPL program as entity linking, we added label name constraints during triple inference, regenerating triples if non-standard label names were produced.  
All the knowledge decomposition data are utilized to simulate handling questions that the knowledge graph uncovered.

\textbf{Knowledge Completion:} We directly instruct the model to undertake knowledge completion tasks, referring to manually written examples.
To fit the entire ReAct agent framework and ensure modularity, we encapsulate the knowledge completion part as a tool, inputting questions and corresponding incomplete knowledge triples to output the mappings of the parameter's knowledge with these triples.

\textbf{Retrieval Augmented Generation Verification:} Since LLM with general instruction tuning and preference alignment are familiar with RAG, we utilize prompt engineering to request the model to generate the correction of knowledge triples based on retrieved relevant documents, incomplete triples with question marks, and corresponding triples with parameter knowledge completion.
We adopt Wikipedia as a retrieval corpus and segment every 256 tokens into a chunk. We build a document index by BM25, searching via concatenated knowledge triples and the origin question and selecting the top ten chunks as external knowledge references.

For the entire ReAct agent framework, we constructed two routes for the agent's planning and calling tools, differentiating whether the necessary knowledge is contained in the knowledge graph.
Utilizing the built parallel training data, we construct two Thought-Action-Observation execution routes of SFT data, considering every scenario elaborated above. 
The agent is tuned using a total of 100k behavior SFT data.

\subsection{System and Use Case}

\begin{figure*}[tp]
  \centering
  \includegraphics[width=0.90\textwidth]{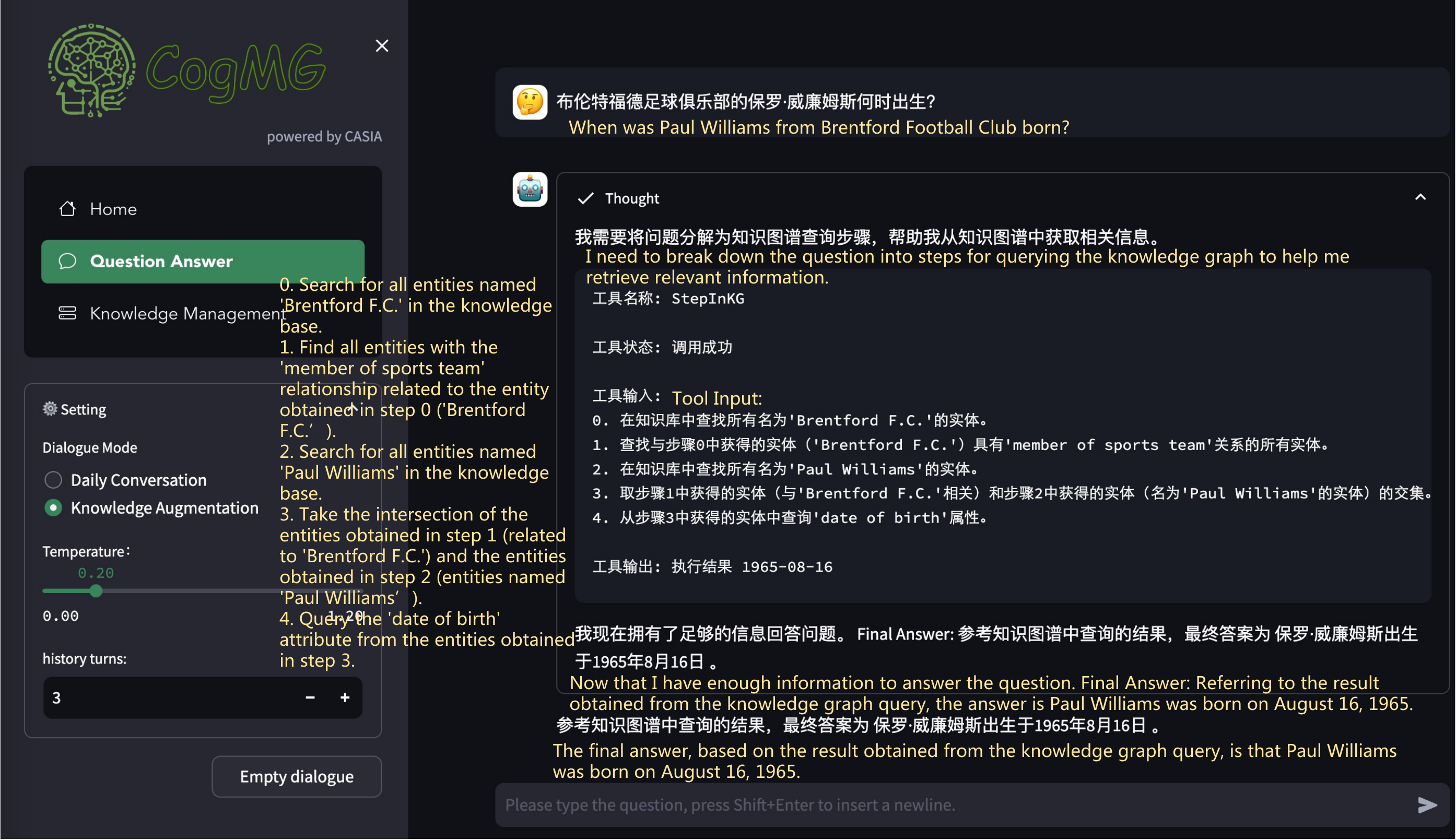}
  \caption{System screenshot.}
  \label{cogmg_ss}
\end{figure*}

\textbf{Knowledge Augmented Generation:} 
Users can input and submit knowledge-intensive questions into the dialogue box at the bottom.
The agent LLM is responsible for dealing with these questions and processes with pre-defined routes.
The Thought-Action-Observation paradigm will be displayed in real time at the corresponding dropdown tab. 
When the knowledge graph cannot support the question-answering process, the model decomposes knowledge and invokes itself for knowledge completion before providing a final answer, as shown in the left of Figure~\ref{figure_case_0}.
Meanwhile, these knowledge triples are recorded in the database.

\textbf{Knowledge Management:} 
In the Knowledge Management section of our system, we design an interactive interface to display all pending instances of knowledge that are not yet covered by the knowledge graph. 
The interface presents the origin of the query that highlighted the knowledge gap, the specific knowledge that is missing, and the results of the model's attempt to complete this knowledge based on its internal parameters.
Administrators can (1) directly integrate this newly completed knowledge into the knowledge graph or opt for (2) further verification through RAG. 
A dropdown tab within the interface provides access to related documents and the outcomes of modifications, facilitating a rigorous validation process. 
Once the verification is complete and any necessary adjustments are made, administrators can seamlessly add the refined knowledge to the graph.
This process not only ensures the continuous expansion and refinement of the knowledge graph but also leverages the administrators' expertise to validate the model-generated knowledge. 
By integrating these human-in-the-loop verification steps, our system enhances the reliability and accuracy of the knowledge graph, making it a more robust resource for answering real-world questions.

\subsection{Experiments}
We further designed and conducted experiments to demonstrate the effectiveness of the CogMG framework.
Sampling questions from the KQA Pro dataset, we tested the following scenarios: (1) \textbf{Direct Answer}: answering using only the backbone LLM without utilizing the knowledge graph; (2) \textbf{CogMG w/o Knowledge}: deleting relevant knowledge from the graph and answering using parameter completion of knowledge; (3) \textbf{CogMG Update}: updating all relevant knowledge, utilizing the graph query results for answering.
Due to the difficulty of exact matching in reflecting the correctness of real answers, we manually evaluated the correctness of 50 questions.
Table~\ref{tab_usecase} illustrates the accuracy under these three scenarios.
Experimental results indicate that directly answering questions using LLM results in lower accuracy due to the lack of precise factual knowledge.
Besides, leveraging the model's knowledge clarification and completion can alleviate some hallucinations and improve accuracy.
Finally, the accuracy of subsequent inquiries is improved after utilizing the collaborative augmentation framework to update the knowledge graph.

\begin{table}[tp]
\centering
\begin{tabular}{lcc}
\hline
\textbf{Method} & \textbf{Accuracy} \\
\hline
Direct Answer & 40\%   \\
CogMG w/o Knowledge & 44\%  \\
CogMG Update & 86\% \\

\hline
\end{tabular}
\caption{Comparison results of the accuracy of question answering in three different scenarios.}
\label{tab_usecase}
\end{table}

\section{Related Work}

\subsection{Knowledge Base Question Answering}

Knowledge Base Question Answering (KBQA) aims to provide answers to natural language questions using Knowledge Bases (KBs) as their primary source of information \citep{bordes2015large,lan2019knowledge}.
Semantic parsing plays a crucial role by mapping questions to a formal language \citep{yih2016value, cai2013semantic}, enabling precise queries on knowledge graphs \citep{bollacker2008freebase, vrandevcic2014wikidata}.
This task format can be regarded as a Seq2Seq paradigm, where formal language sequences are generated based on input question sequences.
From RNN \citep{dong2016language} to BART \citep{cao2022kqa} and GPT \citep{luo2023chatkbqa}, the accuracy of formal language increases gradually with the capability of generative models.
Besides end-to-end generation, \citealp{chen2021retrack} suggested first identifying the entities and schema involved in the problem separately and then utilizing the transducer to generate logical expressions, ensuring the accuracy of logical syntax. Finally, it employs a checker to enhance the semantic consistency of the logical form. 
With the assistance of LLM, KB-BINDER \citep{li2023few} generates a draft of logical expressions using codex and then matches executable programs based on BM25 scores. Thanks to in-context learning, the process can be accomplished with just a few annotated examples. 
Moreover, LLMs' reasoning and planning capabilities can also serve as better assistants for utilizing knowledge graphs without additional training \citep{jiang2023structgpt, sun2023think, jiang2024kg, liu2024enhancing}.
However, these works mainly focus on answering questions within the confines of a given dataset without addressing scenarios where the knowledge graph lacks the necessary information for the question. 
The agent framework \citep{yao2022react, qin2023toolllm, liu2023bolaa} allows for the autonomous selection of alternative tools when faced with knowledge graph misses by design.
However, it does not utilize these gaps as opportunities to enhance the knowledge graph. 

In summary, existing research either overlooks the issue of insufficient knowledge graph coverage or fails to use these deficiencies to improve knowledge graphs actively.

\subsection{Updating Knowledge Graph}

Information extraction concentrates on extracting triples from a wide range of unstructured texts to augment knowledge graphs with new knowledge.
Subject to the model capabilities, the target needs to be split into serval sub-tasks. 
Named entities need to be identified in the text \citep{lample2016neural, yu2020named, qu2023distantly}, followed by the classification of relationships among these entities \citep{miwa2016end, peng2020learning, cheng2021hacred}.
OpenIE \citep{etzioni2008open, stanovsky2018supervised, kolluru2020openie6}, on the other hand, identifies subject-predicate-object triples in one go without being limited by pre-defined schemas in the knowledge graph.
LLMs have unified various information extraction tasks, allowing a single model to generalize across all sub-tasks with supervised fine-tuning or a few examples as a demonstration \citep{lu2022unified, lou2023universal, wang2023instructuie, zhu2023mirror}.
On the other hand, knowledge graph completion \citep{zhang2023making, jiang2023text} through reasoning over existing knowledge can augment the graph by establishing connections between existing nodes.

Our advocated approach of active knowledge updating is more targeted and complements large-scale knowledge updates without contradiction, providing a supplementary mechanism.

\section{Conclustion}

We address two relatively overlooked issues in integrating Large Language Models (LLMs) and Knowledge Graphs (KGs): Incomplete Knowledge Coverage and Knowledge Update Misalignment. 
In response to these challenges, we introduce CogMG, a framework for the collaborative enhancement of LLMs and KGs. 
CogMG tackles the problem of answering questions with knowledge not covered in the graph by explicitly defining and completing relevant knowledge. 
Additionally, it actively collects and verifies knowledge requirements to update the graph. 
Furthermore, we fine-tune an LLM based on an agent framework to implement CogMG and develop a user-friendly interactive system to visualize its capabilities. 
Use cases and experimental results demonstrate the effectiveness of CogMG.

\section{Limitations}

In enhancing large language models with knowledge graphs, we do not introduce more complex and advanced methods such as planning, reasoning, and interaction. We believe that the application of these methods can further improve the effectiveness of the CogMG framework.

On the other hand, actively acquiring updated triples in the real world and automatically incorporating this knowledge into the knowledge graph without human intervention remains challenging. The operation and management of knowledge graphs by large language models are directions for our future work.

\section{Acknowledgments}
This work is supported by the Strategic Priority Research Program of Chinese Academy of Sciences (No. XDA27020203), the National Natural Science Foundation of China (No. 62176257).



\end{document}